# scientific reports

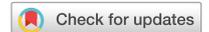

OPEN

# COVID-Net Biochem: an explainability-driven framework to building machine learning models for predicting survival and kidney injury of COVID-19 patients from clinical and biochemistry data

Hossein Aboutalebi[1,3 ✉], Maya Pavlova[2], Mohammad Javad Shafiee[2,3,4], Adrian Florea[5], Andrew Hryniowski[2,4] & Alexander Wong[1,2,3,4]

Since the World Health Organization declared COVID-19 a pandemic in 2020, the global community has faced ongoing challenges in controlling and mitigating the transmission of the SARS-CoV-2 virus, as well as its evolving subvariants and recombinants. A significant challenge during the pandemic has not only been the accurate detection of positive cases but also the efficient prediction of risks associated with complications and patient survival probabilities. These tasks entail considerable clinical resource allocation and attention. In this study, we introduce COVID-Net Biochem, a versatile and explainable framework for constructing machine learning models. We apply this framework to predict COVID-19 patient survival and the likelihood of developing Acute Kidney Injury during hospitalization, utilizing clinical and biochemical data in a transparent, systematic approach. The proposed approach advances machine learning model design by seamlessly integrating domain expertise with explainability tools, enabling model decisions to be based on key biomarkers. This fosters a more transparent and interpretable decision-making process made by machines specifically for medical applications. More specifically, the framework comprises two phases: In the first phase, referred to as the "clinician-guided design" phase, the dataset is preprocessed using explainable AI and domain expert input. To better demonstrate this phase, we prepared a benchmark dataset of carefully curated clinical and biochemical markers based on clinician assessments for survival and kidney injury prediction in COVID-19 patients. This dataset was selected from a patient cohort of 1366 individuals at Stony Brook University. Moreover, we designed and trained a diverse collection of machine learning models, encompassing gradient-based boosting tree architectures and deep transformer architectures, specifically for survival and kidney injury prediction based on the selected markers. In the second phase, called the "explainability-driven design refinement" phase, the proposed framework employs explainability methods to not only gain a deeper understanding of each model's decision-making process but also to identify the overall impact of individual clinical and biochemical markers for bias identification. In this context, we used the models constructed in the previous phase for the prediction task and analyzed the explainability outcomes alongside a clinician with over 8 years of experience to gain a deeper understanding of the clinical validity of the decisions made. The explainability-driven insights obtained, in conjunction with the associated clinical feedback, are then utilized to guide and refine the training policies and architectural design iteratively. This process aims to enhance not only the prediction performance but also the clinical validity and trustworthiness of the final machine learning models. Employing the proposed explainability-driven framework, we attained 93.55% accuracy in survival prediction and 88.05% accuracy in predicting kidney injury complications. The models have been made available through an open-source platform. Although not a production-ready solution, this study aims to serve as a catalyst for clinical scientists, machine learning researchers, and citizen scientists to develop innovative and trustworthy clinical decision support solutions, ultimately assisting clinicians worldwide in managing pandemic outcomes.

[1]Cheriton School of Computer Science, University of Waterloo, Waterloo, Canada. [2]Department of Systems Design Engineering, University of Waterloo, Waterloo, Canada. [3]Waterloo Artificial Intelligence Institute, University of Waterloo, Waterloo, Canada. [4]DarwinAI Corp., Waterloo, Canada. [5]Department of Emergency Medicine, McGill University, Montreal, Canada. ✉email: haboutal@uwaterloo.ca

The COVID-19 global pandemic, with its vast number of infected patients, has placed immense strain on healthcare systems worldwide. This pressure has been exacerbated by the emergence of new COVID variants, leading to increased infection rates and subsequent death tolls[1]. In addition to viral symptoms, recent studies[2,3] have indicated that patients may experience severe kidney complications, such as Acute Kidney Injury (AKI), due to COVID-19 infections. Recognizing these indirect complications and implementing preemptive measures during treatment can significantly enhance a patient's chances of survival[4] and reduce overall healthcare costs[5]. However, the limitations of healthcare resources make it challenging to proactively treat every patient, highlighting the need for effective patient triaging to facilitate prompt, patient-specific interventions.

In this paper, we address the challenge of developing a transparent, explainable model that enables clinicians to guide the design of machine learning models by providing explainability tools that emphasize key biomarkers in the decision-making process. We present COVID-Net Biochem, an explainability-centric framework for constructing machine learning models that predict a patient's survival probability and risk of Acute Kidney Injury (AKI) during hospitalization using clinical and biochemical data in a transparent, systematic manner.

COVID-Net Biochem is a two-stage framework that incorporates both clinician assessments and model insights, obtained via quantitative explainability methods, to facilitate a more profound understanding of the model's decision-making process and the relative influence of various clinical and biochemical markers. This approach enables the development of high-performance, reliable, and clinically relevant machine learning models by iteratively guiding architecture design and training policies based on explainability insights on input markers.

The framework's output includes a diverse array of high-performance machine learning models, encompassing gradient-based boosting tree architectures and deep transformer architectures specifically designed to predict survival probability and kidney injury, along with their dominant clinical and biochemical markers utilized during the decision-making process. Furthermore, our proposed framework can be extended to other healthcare domains.

The explainability insight derived from the model's decision-making process establishes a transparent auditing framework for model decisions. We demonstrate that this capability can be employed in conjunction with new, powerful insights into potential clinical and biochemical markers relevant to prediction outcomes. Consequently, the proposed explainability-driven computer-aided diagnostic framework can support physicians in delivering effective and efficient patient prognoses by providing supplementary outcome predictions based on an extensive array of clinical and biochemical markers, as well as emphasizing key markers pertinent to the task.

Specifically, we utilize explainable AI in the architecture design and training process to ensure the final model's decision-making aligns with clinical perspectives. Several studies, such as[6,7], and[8], have employed explainability toolkits like GSInquire[9] to validate their models' decision-making behaviors in collaboration with clinicians. However, our proposed two-phase model-building framework employs GSInquire to guide the refinement of both data and model design through an iterative clinician-in-the-loop approach, effectively incorporating explainability as an integral part of the model development process, rather than a final validation step.

Importantly, this closed-loop approach is adaptable to any explainability algorithm; thus, the primary contribution lies in the model development framework, rather than the specific use of GSInquire. For the first time, we have explicitly delineated the phases of model building and dataset refinement in which clinicians can participate, and how their guidance can be leveraged to create more reliable machine learning models. This strategy mitigates the risk of bias and inconsistent model behavior, including the use of irrelevant or biased markers in the decision-making process.

### Contributions
The contributions of the proposed work is as follows:

- Introducing a novel explainability-driven framework for data preprocessing and machine learning model design
- Integrating domain expert knowledge (clinicians in our case study) to refine the decision-making process of machine learning models by excluding clinically irrelevant high-impact biomarkers obtained from the explainability model
- Curating a benchmark dataset for COVID-19 patient survival and Acute Kidney Injury (AKI) prediction
- Developing high-accuracy machine learning models for COVID-19 patient survival and Acute Kidney Injury (AKI) prediction

### Motivation
A key generalizable insight we wish to surface in this work is a strategy to counter the largely 'black box' nature of model design at the current state of machine learning in the context of healthcare. Strategies for transparent design are not only critical but very beneficial for building reliable, clinically relevant models in a trustworthy manner for widespread adoption in healthcare. More specifically, while significant advances have been made in machine learning, particularly with the introduction of deep learning, much of the design methodologies leveraged in the field rely solely on a small set of performance metrics (e.g., accuracy, sensitivity, specificity, etc.) to evaluate and guide the design process of the models. Such 'black box' design methodologies provide little insight into the decision-making process of the resulting machine learning models, and as such even the designers themselves have few means to guide their design decisions in a clear and transparent manner. This is particularly problematic given the mission-critical nature of clinical decision support in healthcare and can lead to a significant lack of trust and understanding by clinicians in computer-aided diagnostics. Furthermore, the lack of interpretability creates significant accountability and governance issues, particularly if decisions and recommendations made by machine learning models result in negative patient impact.

Motivated to tackle the challenges associated with 'black box' model design for clinical decision support, in this work we propose an explainability-driven development framework for machine learning models that can be extended to multiple healthcare domains such as COVID-19 survival and acute kidney injury prediction. The framework provides a two-phase approach in which a diverse set of machine learning models are designed and trained on a curated dataset and then validated using both an quantitative explainability technique to identify key features as well as a manual, qualitative clinician validation of the highlighted features. The second phase consists of leveraging the explainability-driven insights to revise the data and design of the models to ensure high classification performance from relevant clinical features. The resulting outputs from the development process are high-performing, transparent detection models that not only provide supplementary outcome predictions for clinicians but also quantitatively highlight important factors that could provide new insights beyond standard clinical practices.

### Related works

Since the beginning of the COVID-19 pandemic, there has been a significant global emphasis on enhancing effective screening methods. Accurate and efficient patient screening is crucial for providing timely treatment and implementing isolation precautions. Consequently, numerous research efforts have been directed towards employing deep learning models for the automatic screening of COVID-19 patients. Studies have demonstrated that deep learning can facilitate the diagnosis of COVID-19 cases based on Chest X-ray (CXR) images with acceptable accuracy[10–15]. Additionally, other works have utilized Computed Tomography (CT) images for diagnosing COVID-19 cases[16–18]. Moreover, various approaches have been suggested for assessing the severity of COVID-19 patients using medical imaging modalities[19–21].

The application of computer-aided diagnostics for screening medical images of COVID-19 patients has gained significant traction. However, limited research has focused on utilizing machine learning models for assessing patient survival and the development of Acute Kidney Injury (AKI). A major drawback of the algorithms proposed thus far is their lack of interpretability in model design, which impedes their adoption by clinicians in real-world settings. The provision of interpretable results is crucial, as it aids clinicians in validating the decision-making process of the prediction model. Although some studies[6–8] have incorporated explainability mechanisms, their approaches primarily concentrate on final model validation. These studies do not capitalize on an iterative clinician-in-the-loop data and model development framework, which could enhance their overall effectiveness.

Table 1 shows some of the related works (including our work) for survival prediction and provides the pros and cons of each model. In this regards, Spooner et al.[22] investigates the performance and stability of ten machine learning algorithms, paired with different feature selection methods, for predicting the time to dementia onset in patients. The analysis employs high-dimensional, heterogeneous clinical data to evaluate the efficacy of these machine-learning models in conducting survival analysis. The work of Nemati et al. explores the application of statistical models and machine learning techniques to real-world COVID-19 data, aiming to predict patient discharge time and assess the influence of clinical information on hospital length of stay. In the study by Brochers[23], a multi-omic machine learning model was developed, incorporating the concentrations of 10 proteins and five metabolites to predict patient survival outcomes. In addition, in Taheriyan et al.[24], the authors have developed a deep learning-based survival prediction model utilizing demographic and laboratory data to forecast patient outcomes.

| Related works | | | |
|---|---|---|---|
| Study | Pros | Cons | Year |
| A comparison of machine learning methods for survival analysis of high-dimensional clinical data for dementia prediction[22] | Studies different machine learning models<br>Identify high impact biomarker | The code is not open-sourced<br>The models are not explainable | 2020 |
| Machine-Learning Approaches in COVID-19 Survival Analysis and Discharge-Time Likelihood Prediction Using Clinical Data[27] | The code is open-sourced<br>Studies different machine learning models | The high impact biomarkers are not identified<br>The models are not explainable<br>Models accuracy are not high | 2020 |
| Early prediction of COVID-19 patient survival by targeted plasma multi-omics and machine learning[23] | Using the main biomarker in the design of the predictive model<br>High accuracy model | The models are not explainable<br>The code is not open-sourced<br>Lacks using other machine learning models and comparing them | 2021 |
| Prediction of COVID-19 Patients' Survival by Deep Learning Approaches[24] | The design process of the model is explained<br>High accuracy model | The models are not explainable<br>The code is not open-sourced<br>Lacks using other machine learning models and comparing them | 2022 |
| COVID-Net Biochem (our work) | Use explainable models<br>High accuracy model<br>Code is available<br>Studies different machine learning and deep learning | The training process has more stages and takes longer to train<br>Requires domain expert knowledge in the design of the predictive models | 2023 |

**Table 1.** Comparison of related works for survival prediction.

In a study closely related to our proposed method, Gladding et al.[25] introduced an approach that employs a machine learning model to diagnose COVID-19 and other diseases using hematological data. Additionally, Erdi et al.[26] developed a novel deep learning architecture to detect COVID-19 based on laboratory results. While these studies primarily concentrate on identifying COVID-19 positive cases, our research aims to predict patient survival and the risk of developing AKI during hospitalization by analyzing biochemical data. We propose an end-to-end transparent model development framework, which can be adapted for use in other healthcare domains.

## Explainability-driven framework methodology for building machine learning models for clinical decision support

In this section, we present a comprehensive framework designed to construct high-performance, clinically-robust machine learning models utilizing clinical and biochemical markers for transparent and reliable detection. We demonstrate the effectiveness of this framework by employing clinical and biochemical markers to develop machine learning models for predicting acute kidney injury (AKI) and survival outcomes in COVID-19 patients. The proposed COVID-Net Biochem framework, depicted in Fig. 1, consists of two primary phases:

### Clinician-guided design phase

The first phase starts with the preparation of a benchmark dataset of carefully curated clinical and biochemical markers based on clinical assessment. While a plethora of markers may be collected for a patient cohort, only a selected number of markers are relevant for a given predictive task while others may not only be irrelevant but also misleading when leveraged. Therefore, in this phase, we remove clinically irrelevant markers through consultations with clinicians who have domain knowledge for the given task. Next, a collection of different machine learning models with a diversity of gradient-based boosting tree architectures and deep transformer architectures are designed and trained on the constructed benchmark dataset.

### Explainability-driven design refinement phase

The second phase begins with explainability-driven validation of model performance and behavior to gain a deeper understanding of the model's decision-making process and to acquire quantitative insights into the influence of clinical and biochemical markers. In this paper, we employ the computational explainability technique called GSInquire[9] to conduct this evaluation. Next, we analyze and interpret the decision-making process of the model through the identified relevant predictive markers and use the insights iteratively to develop progressively better and more clinically relevant machine learning models. More specifically, if all of the clinical and biochemical markers identified are driving the decision-making process of a given model and these markers are verified to be clinically sound based on the clinical assessment of the explainability results, the model is accepted as the final model. Otherwise, we return to the first phase where the irrelevant markers are discarded for that given model, and a new model architecture is trained and produced with hyperparameter optimization and then tested again for phase 2. This iterative approach not only removes the influence of quantitatively and clinically irrelevant clinical and biochemical markers but also eliminates the markers that may dominate the decision-making process when they are insufficient for clinically sound decisions (e.g., the heart rate clinical marker may be clinically relevant when used with other markers but should not be solely relied upon for survival prediction from COVID-19 due to its general severity implication). This iterative process continues until the model heavily utilizes only clinically sound input markers to great effect in its decision-making process.

In this particular study, the initial clinician-guided design phase consists of constructing a new, clinician curated benchmark dataset of clinical and biochemical data from a patient cohort of 1366 patients at Stony Brook University[28]. Next, a collection of models with the following architectures were trained on the constructed benchmark dataset: (i) TabNet[29], (ii) TabTransformer[30], (iii) FTTransformer[31], (iv) XGBoost[32], (v) LightGBM[33], and (vi) CatBoost[34]. TabNet focuses on employing sequential attention to score features for decision-making, ultimately improving the model interpretability compared to previously proposed deep learning models for tabular datasets[29]. Furthermore, TabTransformer and FTTransformer models utilize a more recent transformer architecture designed to process tabular datasets. In practice, transformer models have shown higher performance on most well-known datasets[30,31,35] and are thus leveraged for this particular patient detection task. The gradient boosting algorithms of LightGBM[33] and CatBoost[34] are also utilized as they rely on creating and learning an ensemble of weak prediction models (decision trees) by minimizing an arbitrary differentiable loss function. In addition, for a baseline comparison of performance, both Logistic Regression and Random Forest models are added to the results (Fig. 2).

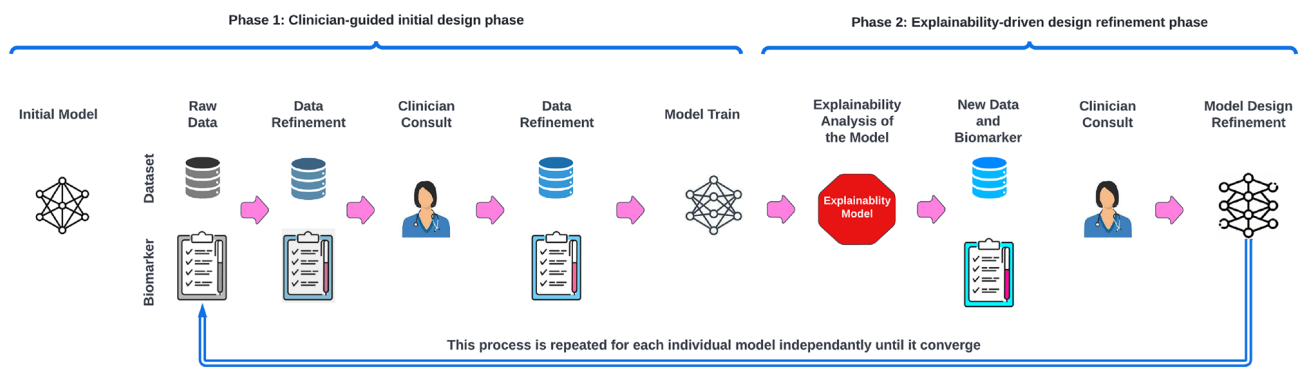

**Figure 1.** Overview of the proposed explainability-driven framework for building machine learning models for clinical decision support.

During the explainability-driven refinement phase, we perform quantitative validation of the model's performance and behavior using GSInquire[36]. GSInquire is a cutting-edge explainability method proven to generate explanations that more accurately represent the decision-making process compared to other prominent techniques in the literature. This approach allows for the allocation of influence values to each clinical and biochemical marker, illustrating their impact on the model's predictions. Lastly, a clinical evaluation of the explainability-driven insights was carried out by an experienced clinician with over 8 years of expertise.

### Data preparation and refinement

In this section, we provide a comprehensive overview of the data preparation process used in constructing the benchmark dataset for COVID-19 patient survival and AKI prediction, as well as the clinical and biochemical marker selection process conducted based on explainability-driven insights in the design refinement phase. The proposed dataset is built by carefully selecting clinical and biochemical markers based on clinical assessment from a patient cohort curated by Stony Brook University[28]. More specifically, the clinical and biochemical markers were collected from a patient cohort of 1336 COVID-19 positive patients and consists of both categorical and numerical markers. The markers are derived from patient diagnosis information, laboratory test results, intubation status, oral temperature, symptoms at admission, as well as a set of derived biochemical markers from blood work. Table 2 demonstrates the numeric clinical and biochemical markers from the patient cohort and their associated dynamic ranges.

The categorical clinical features consist of *"gender"*, *"last status"* (discharged or deceased), *"age"*, *"is ICU"* (admitted to ICU or not), *"was ventilated"* (received ventilator or not), *"AKI during hospitalization"* (true or false), *"type of therapeutic received"*, *"diarrhea"*, *"vomiting"*, *"nausea"*, *"cough"*, *"was antibiotic received"* (true or false), *"other lung diseases"*, *"urine protein"*, *"smoking status"*, and *"abdominal pain"*.

### Target value

In this study, the patient's *"last status"* is used as the target value for predicting the COVID-19 survival chance given the patient's symptoms and status. The onset of *"AKI during hospitalization"* is leveraged as the target value for the task of predicting the development of kidney injury during hospitalization for COVID-19. Figure 3 demonstrates the distribution of these two target values respectively in the final curated benchmark dataset for the leveraged patient cohort. As a result of the high negative to positive imbalances in the final dataset, a per batch upsampling of positive patients was performed during model training as a batch re-balancing technique.

### Missing values and input transformations

Substantial missing values in an individual marker were handled by removing from the benchmark dataset all markers that had more than 75% of samples with a missing value. To replace missing values within the resultant markers, we followed the strategy introduced in TabTransformer[30] where the missing value is treated as an

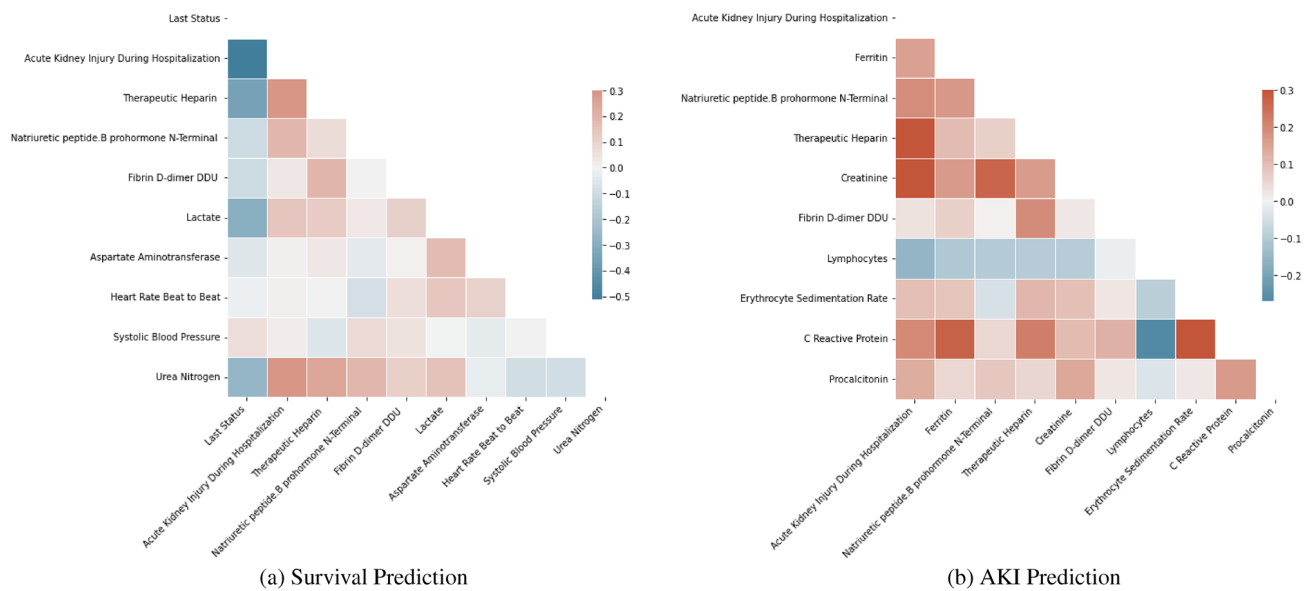

**Figure 2.** Pearson correlation coefficients between key identified clinical and biochemical markers for COVID-19 patient survival and AKI prediction.

| Clinical/biochemical markers (numeric) | Minimum value | Maximum value |
|---|---|---|
| Invasive ventilation days | 0 | 40 |
| Length of stay | 1 | 96 |
| Oral temperature | 34 | 39.8 |
| Oxygen saturation in arterial blood by pulse | 55 | 100 |
| Respiratory rate | 11.0 | 95 |
| Heart rate beat by EKG | 6 | 245 |
| Systolic blood pressure | 55 | 222 |
| Mean blood pressure by non invasive | 40 | 168 |
| Neutrophils in blood by automated count | 0.36 | 100 |
| Lymphocytes in blood by automated count | 0.36 | 100 |
| Sodium [moles/volume] in serum or plasma | 100 | 169 |
| Aspartate aminotransferase in serum or plasma | 8 | 2786 |
| Aspartate aminotransferase in serum or plasma | 8 | 2909 |
| Creatine kinase in serum or plasma | 11 | 6139 |
| Lactate in serum or plasma | 5 | 23.8 |
| Troponin T.cardiac in serum or plasma | 0.01 | 1.81 |
| Natriuretic peptide.B prohormone N-terminal in serum or plasma | 5 | 267,600 |
| Procalcitonin in serum or plasma immunoassay | 0.02 | 193.5 |
| Fibrin D-dimer DDU in platelet poor plasma | 150 | 63,670 |
| Ferritin [mass/volume] in serum or plasma | 5.3 | 16,291 |
| Hemoglobin A1c in blood | 4.2 | 17 |
| BMI ratio | 11.95 | 92.8 |
| Potassium [moles/volume] in serum or plasma | 2 | 7.7 |
| Chloride [moles/volume] in serum or plasma | 60 | 134 |
| Bicarbonate [moles/volume] in serum | 6 | 43 |
| Glomerular filtration rate | 2 | 120 |
| Erythrocyte sedimentation rate | 5 | 145 |
| Cholesterol in LDL in serum or plasma | 12 | 399 |
| Cholesterol in VLDL [mass/volume] in serum | 8 | 79 |
| Triglyceride | 10 | 3524 |
| HDL | 10 | 98 |

**Table 2.** Example numerical clinical and biochemical markers collected from the patient cohort.

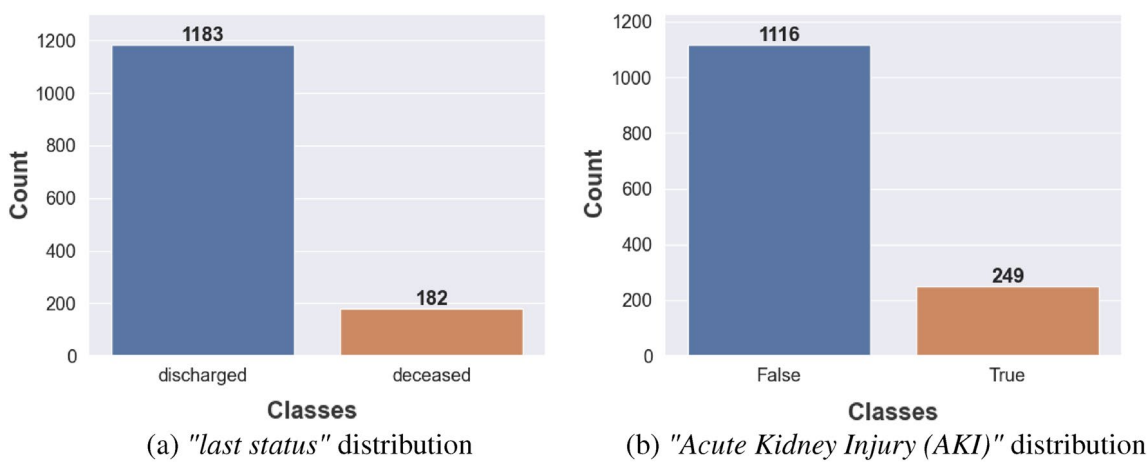

(a) *"last status"* distribution    (b) *"Acute Kidney Injury (AKI)"* distribution

**Figure 3.** Distribution of *"last status"*, *"AKI"*.

additional category. We also found that both transformer models and gradient boosting tree models are resilient against missing values, and replacing the missing value with a constant gives an equally competitive result.

For input preprocessing, we found that using different types of input transformations do not substantially change the final result in our models. In this regard, we examined the MinMax scaler, the uniform transformer, and the normal distribution transformer (all available in sickit-learn's preprocessing method[37]). None of them provided any better results.

### Clinician guided marker selection
To carefully curate the benchmark dataset based on clinical data, we consulted with a clinician with over 8 years of experience. Through the consultation, we identified markers that are clinically irrelevant and may result in biases being learned by the machine learning models. More specifically, confounding factors such as *"heart rate"*, *"invasive ventilation days"* were removed after consulting with the clinician as their impact on survival and AKI prediction were not directly clinically relevant.

### Explainability-driven clinical and biochemical marker refinement
In the explainability-driven design refinement phase, we conduct a quantitative analysis with GSInquire[9] of the decision-making processes of each individually trained model within the collection of model designs to identify the influence of inputted markers. After identifying the most quantitatively important markers to the decision-making processes, we presented these explainability results to the clinician to not only gain valuable clinical insights into the clinical soundness of the machine learning models but also to identify the non-relevant markers among these that the models rely on so that they will be excluded in the next round of model training. As an example, after conducting an explainability-driven assessment on the machine learning models with LightGBM and CatBoost architectures, we observed that the clinical marker *"Length of Stay"* had the highest quantitative impact on the decision-making process of said models for the AKI prediction (see Fig. 6). After clinical consultation, we found out this clinical marker has little clinical value in determining the likelihood of AKI. As a result, in the next phase of model design and training, the *"Length of Stay"* marker was excluded. This process continued until only the relevant markers for our prediction tasks were utilized by the individual models. It is very important to note that explainability-driven assessment was conducted on each model independently, and as such the end result is that each model is uniquely tailored to specific clinical and biochemical markers.

Finally, to better understand the correlation between clinical and biochemical markers, Fig. 2 shows the correlation between the top ten markers for onset AKI during hospitalization and patient's last status target labels. As seen, for the target *"last status"*, AKI has the highest correlation. On the other hand, for the target label AKI during hospitalization, *"urine protein"*, *"Therapeutic Heparin"*, *"Fibrin D Dimer"*, *"Creatinine"* and *"Glomerular Filtration"* have the highest correlation values. The influence of such discussed markers on each target label at the individual model level is discussed further in the following Explainability section.

## Experimental results
In this section, we describe the experimental results and training procedures for the different machine learning models created using the proposed framework for the purpose of predicting COVID-19 patient survival and AKI development in COVID-19 patients during hospitalization. As mentioned earlier, we designed six different machine learning models for the aforementioned prediction tasks using the following architecture design patterns: TabTransformer, TabNet, FTTransformer, XGBoost, LightGBM, and CatBoost. Our training procedure is guided by not only the standard metrics of accuracy, precision, and recall but also by identified explainability results.

### The working environment
We executed our code using Python, specifically leveraging the Scikit-learn[37] and PyTorch Tabular[38] libraries for implementation purposes. The computer hardware employed featured an Intel(R) Xeon CPU, complemented by an NVIDIA RTX 6000 graphics card.

### Experimental configuration
*Data preprocessing*
The preprocessing is to ensure the data is suitably prepared for the machine learning algorithms.

The dataset comprised both categorical and numerical values. For the categorical data, we employed one-hot vector encoding as a transformation technique.

To address missing values in the data, we experimented with various strategies, such as iterative imputer and KNN imputer from Scikit-learn[37]. Nevertheless, we found that substituting all missing values with a consistent constant value yielded superior performance.

*Model configurations*
We did extensive grid search to find the best hyperparemeter for each machine-learning models we used in this study. To select the best hyperparemeter, we used Monte Carlo cross-validation to create 5 random splits of the dataset. The averaged results are reported in the Tables 3, and 5. Here, we discuss the details of our grid search for experiment configuration and finding the optimal parameter for each machine learning model:

- For the LightGBM model[33], the learning rate search space ranged from 0.01 to 1. The optimal learning rates were determined to be 0.2 for survival prediction and 0.09 for AKI prediction. The search space for the number of leaves spanned from 10 to 100, with the best values identified as 16 leaves for survival prediction and 40 leaves for AKI prediction. The maximum tree depth search space was set within the interval [3, 16] with a step size of one, yielding the most favorable depths of 12 for survival prediction and 11 for AKI prediction. Lastly, the lambda value search space ranged between 1 and 15, also with a step size of one. The optimal lambda values were found to be 4 for survival prediction and 13 for AKI prediction.
- For the XGBoost model[32], a grid search was conducted to determine the optimal maximum tree depth within the range of [3, 10] with a step size of one. The best values were 9 for survival prediction and 4 for AKI prediction. The minimum child weight search was performed within the interval [1, 30] using a step size of 5, resulting in the best values of 1 for survival prediction and 16 for AKI prediction. The gamma value search space spanned from 0 to 30 with a step size of 5, yielding the best value of 0 for both survival and AKI predictions. The number of estimators search was conducted within the range of [50, 2000], identifying the optimal values of 500 for survival prediction and 1500 for AKI prediction. Lastly, the learning rate search space ranged from 0.001 to 0.1, with the best values determined as 0.01 for survival prediction and 0.09 for AKI prediction.

| Survival prediction | | | | |
|---|---|---|---|---|
| Model | Accuracy (%) | Precision (%) | Recall (%) | F1 score |
| FTTransformer[31] | 87.17 | 88.28 | 98.39 | 0.9302 |
| TabTransformer[30] | 87.91 | 92.66 | 93.50 | 0.9307 |
| TabNet[40] | 86.66 | 88.91 | 96.70 | 0.9263 |
| XGBoost[32] | 92.30 | **94.92** | 96.28 | 0.9560 |
| LightGBM[33] | 93.55 | 95.52 | 97.13 | 0.9631 |
| CatBoost[34] | **93.55** | 94.20 | 98.64 | **0.9637** |
| Random forest[39] | 87.03 | 87.00 | **100.0** | 0.9305 |
| Logistic regression[37] | 87.32 | 88.86 | 97.63 | 0.9304 |
| XGBoost with[22] tunning | 90.69 | 93.22 | 96.28 | 0.9472 |
| Random forest with[22] tunning | 88.20 | 88.04 | 100.00 | 0.9363 |

**Table 3.** Accuracy, precision, recall, and F1 score of tested models for survival prediction. Significant values are in bold.

- For CatBoost model[34], we conducted a grid search for the number of estimators within the range of [50, 1000], and the best value was determined to be 500. For the learning rate, the search space ranged from 0.001 to 0.1, with the optimal value identified as 0.01. The search space for the depth of the tree spanned from 1 to 10. The best values were found to be 7 for survival prediction and 10 for AKI prediction.
- For the Random Forest model[39], we conducted a grid search for the number of estimators within the range of [1, 100] using a step size of 20. The optimal values were found to be 1 for survival prediction and 61 for AKI prediction. For the maximum depth of the tree, the search space spanned from 1 to 10 with a step size of one. The best values were determined as 10 for AKI prediction and 7 for survival prediction.
- For TabTransformer[30], TabNet[40], and FTTransformer[31], we performed grid searches specifically for learning rate, batch size, and the number of training epochs. We found that the learning rate had a more significant impact than the other parameters, with a search space ranging from 0.0001 to 0.1. For TabNet, the optimal learning rate was 0.001, with the number of epochs set at 150 and a batch size of 128. For FTTransformer, a batch size of 128 and a learning rate of 0.001 yielded higher performance. In terms of AKI prediction, we found that an epoch setting of 200 resulted in higher accuracy, while a setting of 100 was better for survival prediction.

### Survival prediction

For this task, the binary label of *last status*, depicting whether a patient became deceased or discharged during COVID-19 hospitalization, was leveraged as the target ground-truth label. During model development, the dataset was split into 75% for training, 5% for validation, and 20% for testing. For the TabTransformer, TabNet, and FTTransformer model architectures, we used Adam optimizer for all model training. The training procedure was done in parallel with getting explainability results for the model. In this regard, we discarded features "*heart rate*", "*length of stay*", "*invasive ventilation days*" as models tend to heavily rely on these less relevant factors for decision making.

For gradient boosting models XGBoost, CatBoost, and LightGBM, as prescribed in the previous section, we employed a grid search for finding the optimal hyperparameters.

The accuracy, precision, recall, and F1 scores for each model are shown in Table 3. In addition, Table 4 also shows the confusion matrix for CatBoost and TabTransformer. As it is shown, CatBoost had the best overall performance, achieving the highest accuracy of 93.55% on the test set and highest F1 score of 0.9637. Among deep learning models, the TabTransformer had the best performance with an accuracy of 87.91%. Also, both TabTransformer and CatBoost had above 92% results for recall and precision.

| Class | Discharged | Deceased |
|---|---|---|
| Confusion matrix CatBoost | | |
| Discharged | 233 | 4 |
| Deceased | 13 | 23 |
| Confusion matrix LightGBM | | |
| Discharged | 229 | 8 |
| Deceased | 9 | 27 |

**Table 4.** Confusion matrix for CatBoost and LightGBM for survival prediction.

| AKI prediction | | | | |
|---|---|---|---|---|
| Model | Accuracy (%) | Precision (%) | Recall (%) | F1 score |
| FTTransformer[31] | 82.12 | 51.37 | 39.59 | 0.4156 |
| TabTransformer[30] | 84.39 | 56.91 | **61.20** | 0.5890 |
| TabNet[40] | 82.05 | 45.37 | 14.40 | 0.2071 |
| XGBoost[32] | **88.05** | 68.40 | 64.79 | **0.6653** |
| LightGBM[33] | 87.91 | 75.75 | 50.80 | 0.6053 |
| CatBoost[34] | 87.17 | 67.79 | 57.20 | 0.6200 |
| Random forest[39] | 86.15 | **69.79** | 43.20 | 0.5334 |
| Logistic regression | 81.97 | 51.73 | 21.19 | 0.3001 |
| XGBoost with[22] tunning | 83.07 | 54.35 | 42.79 | 0.4777 |
| Random forest with[22] tunning | 83.29 | 58.35 | 31.60 | 0.4090 |

**Table 5.** Accuracy, precision, recall, and F1 score of tested models for AKI prediction. Significant values are in bold.

### AKI prediction
In this task, the binary label "Acute Kidney Injury during hospitalization" was employed as the target ground-truth label, while the input marker "last status" was eliminated due to its irrelevance as a clinical marker. The model training procedures utilized for this task mirrored those previously outlined for the survival prediction task.

The results for each model are depicted in Table 5, with a confusion matrix for LightGBM and TabTransformer shown in Table 6. As it can be seen, the XGBoost model had the best overall performance achieving the highest accuracy of 88.05% on the test set.

The benchmark dataset created in this study and the link to the code is available here (https://github.com/h-aboutalebi/CovidBiochem).

### Comparison with other works
In our research, we conducted a comparative analysis with the study presented by[22], which predicts dementia survival using machine learning models. This study is the most closely related to our own work, as it also employs machine learning techniques for survival prediction and utilizes tabular datasets. Unfortunately, we were unable to access the original source code for their implementation, so we re-implemented their XGBoost and Random Forest models based on the preprocessing steps outlined in their paper. The results can be found in Tables 3 and 5.

To ensure a faithful replication of[22], we employed the imputation techniques described in their paper to address missing data and used one-hot encoding for categorical data transformation. We observed that their reported results indicated higher accuracy when all features were included. Thus, we retained the same set of features for a fair comparison between our work and theirs. As shown in Tables 3 and 5, our proposed tuning approach outperformed[22] in all cases, except for the Random Forest model used in survival prediction.

### Explainability results
As explained earlier, the trained models from phase one of the development framework were then audited via explainability-driven performance validation to gain insights into their decision-making process to inform the design refinement in phase two of the process. We leveraged GSInquire[9] to provide the quantitative influence of input clinical and biochemical markers. More specifically, GSInquire provides impact scores for each marker based on their influence on the outcome prediction through an inquisitor within a generator-inquisitor pair. These actionable insights were then further validated by a clinician to ensure clinical relevance, and are later employed by the framework to make design revisions to the models accordingly. Discussed below are the results from the explainability-driven validation for the final, refined models.

| Class | False | True |
|---|---|---|
| Confusion matrix LightGBM | | |
| False | 217 | 6 |
| True | 22 | 28 |
| Confusion matrix XGBoost | | |
| False | 210 | 13 |
| True | 16 | 34 |

**Table 6.** Confusion matrix for LightGBM and XGBoost for AKI prediction.

Figures 4 and 5 display the 10 most influential clinical and biochemical markers relevant to COVID-19 survival and AKI prediction for the top-performing models of XGBoost, LightGBM, and CatBoost, respectively. For predicting COVID-19 patient survival, the marker indicating whether a patient has experienced acute "kidney injury (AKI) during hospitalization" has the highest impact on model predictions, which aligns with our clinician's suggestion. In this context, we observed in Fig. 2 a direct correlation between survival and acute kidney injury Fig. 6.

Interestingly, Fig. 5 reveals that all models consider "Creatinine" as the most critical biomarker for Acute Kidney Injury (AKI) prediction. As the creatinine serum concentration rises from its baseline, it signifies a reduced ability of the kidney to filter toxins and regulate water and salt balance. In a clinical setting, this results in electrolyte imbalances, volume overload, and toxin accumulation, which can necessitate hemodialysis. It is therefore logical that AKI is an essential predictor of mortality, with creatinine being a vital component in measuring the degree of AKI.

### Statistical analysis

In Figs. 7 and 8, we present the Receiver Operating Characteristic (ROC) curves and corresponding Area Under the Curve (AUC) values for XGBoost, LightGBM, and CatBoost algorithms, applied to both survival and Acute Kidney Injury (AKI) prediction tasks. As evident from the figures, predicting AKI proves to be a more challenging task compared to survival prediction for all three models. Nonetheless, each model demonstrates a high AUC (above 0.89), indicating the effectiveness of the training procedures employed.

*Ablation study*

In Tables 7 and 8, we present the results of an ablation study, which involves removing the highest impact biomarker from the models. For survival prediction, based on the information from the explainability graphs in the previous section, we removed "Acute Kidney Injury during hospitalization," while for AKI prediction, we removed "Creatinine." It is evident that the performance of almost all models has decreased, with a few exceptions, such as Logistic Regression and Random Forest. This demonstrates that the explainability model was successful not only in identifying the highest-impact biomarker but also in pinpointing the biomarkers that significantly affect the models' performance.

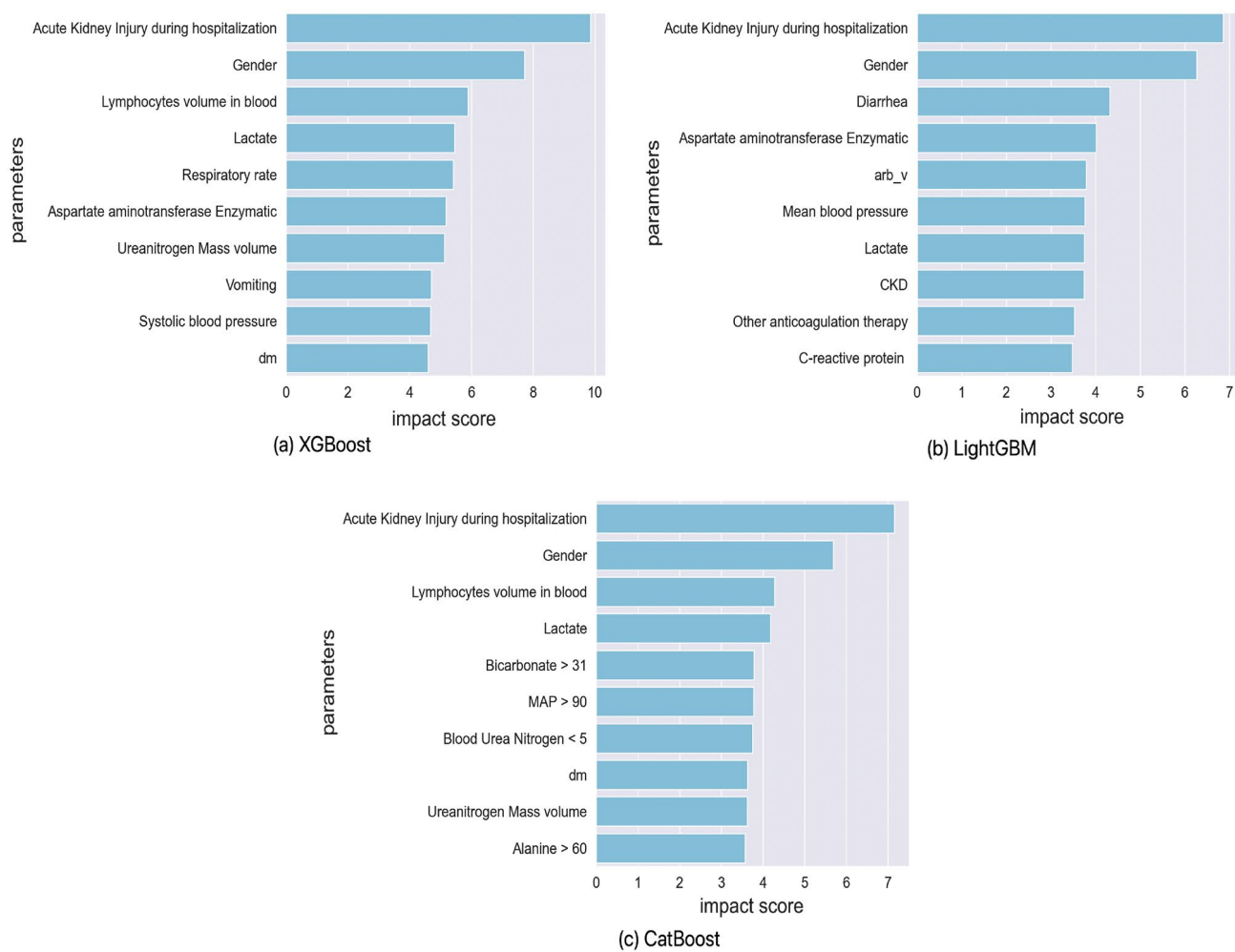

**Figure 4.** Top 10 markers identified through explainability-performance validation for XGBoost, LightGBM, and CatBoost survival prediction models.

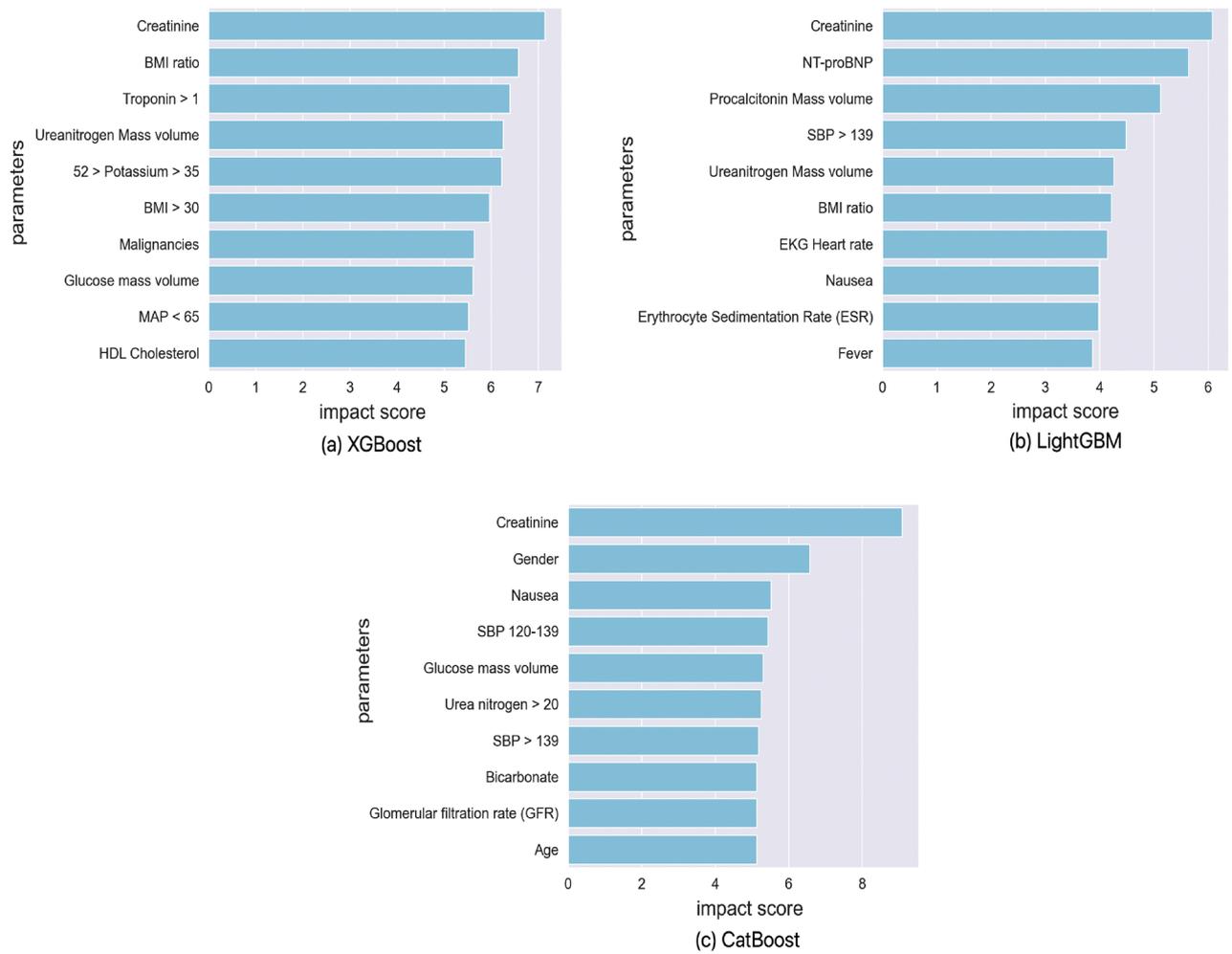

**Figure 5.** Top 10 markers identified through explainability-performance validation for XGBoost, LightGBM, and CatBoost AKI prediction models.

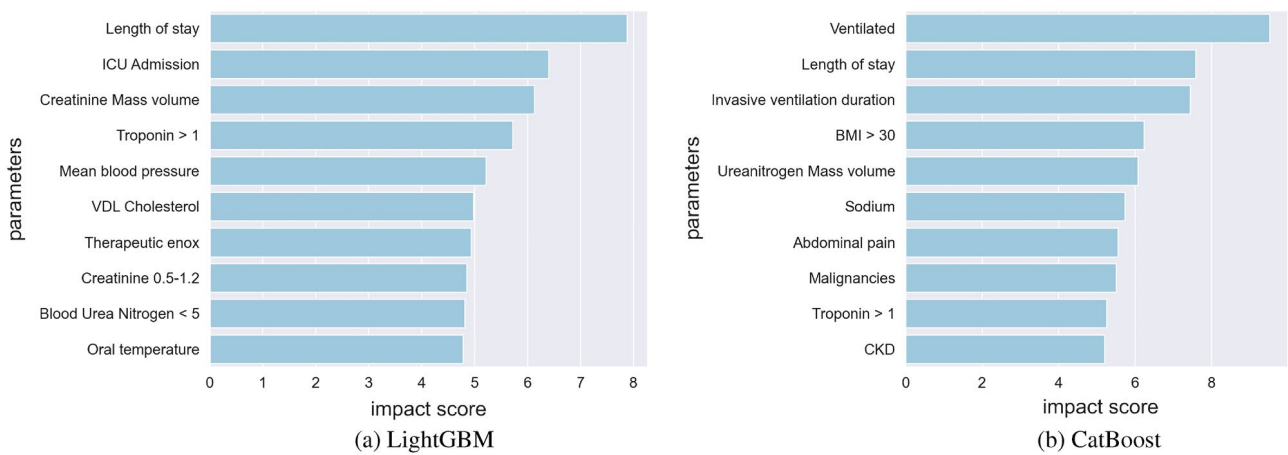

**Figure 6.** Top 10 clinical and biochemical markers identified through explainability-performance validation for LightGBM and CatBoost models for AKI prediction with the inclusion of the length of stay parameter in available patient data.

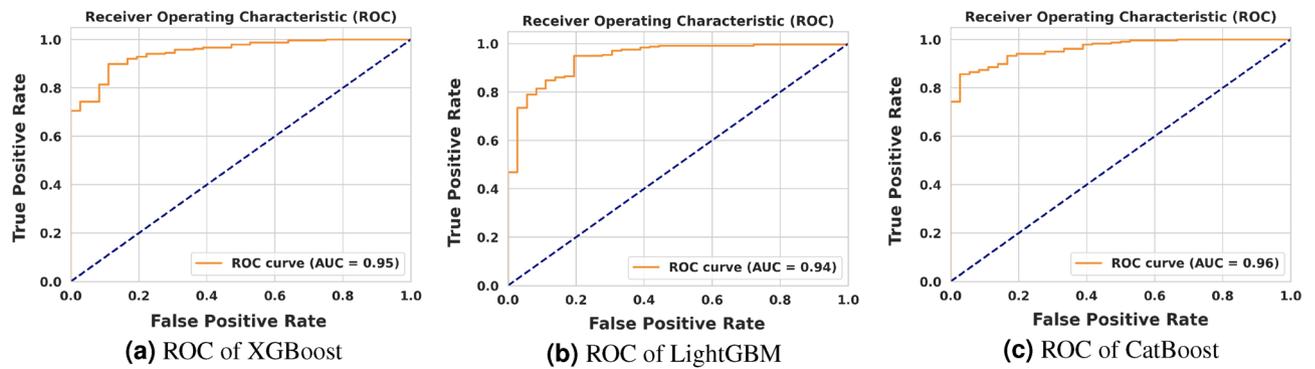

**Figure 7.** ROC of XGBoost, LightGBM, and CatBoost on survival prediction.

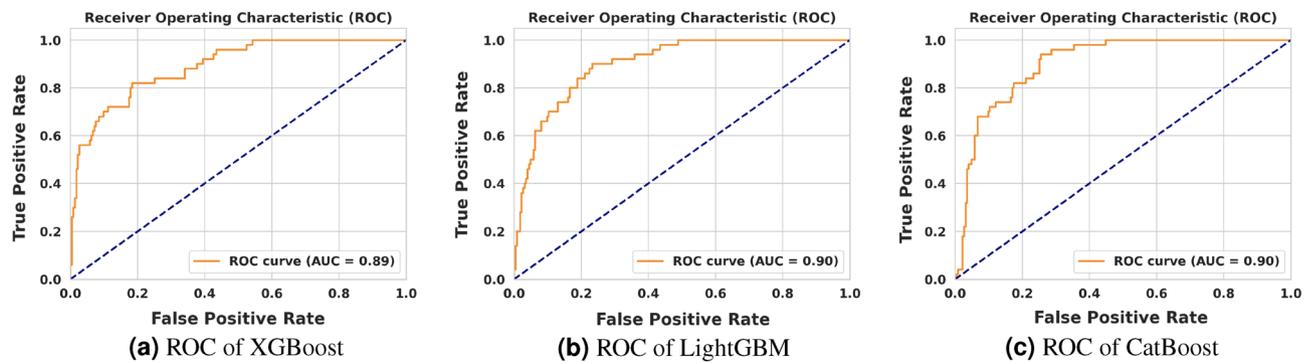

**Figure 8.** ROC of XGBoost, LightGBM, and CatBoost on AKI prediction.

| AKI prediction | | | | |
|---|---|---|---|---|
| Model | Accuracy (%) | Precision (%) | Recall (%) | F1 score |
| XGBoost[32] | 87.17 | 66.08 | 61.60 | 0.6374 |
| LightGBM[33] | 87.32 | 71.77 | 50.80 | 0.5943 |
| CatBoost[34] | 86.52 | 66.84 | 52.40 | 0.5873 |
| Random forest[39] | 84.98 | 67.14 | 34.80 | 0.4571 |
| Logistic regression | 82.05 | 52.33 | 22.80 | 0.3175 |

**Table 7.** Accuracy, precision, recall, and F1 score of tested models for AKI Prediction with removing creatinine.

| Survival prediction | | | | |
|---|---|---|---|---|
| Model | Accuracy (%) | Precision (%) | Recall (%) | F1 score |
| XGBoost[32] | 92.23 | 94.84 | 96.28 | 0.9556 |
| LightGBM[33] | 91.13 | 92.35 | 97.89 | 0.9504 |
| CatBoost[34] | 92.45 | 93.21 | 98.48 | 0.9577 |
| Random forest[39] | 89.45 | 91.15 | 97.29 | 0.9412 |
| Logistic regression | 87.69 | 89.15 | 97.72 | 0.9323 |

**Table 8.** Accuracy, precision, recall, and F1 score of tested models for survival prediction when removing acute kidney Injury biomarker.

## Conclusions
In this work, we presented an explainability-driven framework for developing transparent machine learning models that utilize clinically relevant markers for prediction. As a proof of concept, we applied this framework to predict survival and acute kidney injury during the hospitalization of COVID-19 patients. Experimental results demonstrate that the constructed machine learning models were not only able to achieve high predictive performance but also relied on clinically-sound clinical and biochemical markers in their decision-making processes. In this context, we provided a thorough evaluation of the models' accuracy, recall, precision, F1 score, and confusion matrix using a benchmark dataset. Additionally, we interpreted the models' decision-making process employing quantitative explainability techniques via GSInquire. Ultimately, we revealed that the model considers acute kidney injury as the primary factor in determining the survival likelihood of COVID-19 patients and relies on Creatinine biochemical markers as the principal factor for assessing the risk of developing kidney injury, which aligns with clinical interpretation.

## Limitations
Our results showcase the potential of developing more explainable models to tackle healthcare challenges. However, it is crucial to conduct further experiments to validate the findings from this study for a broader range of clinical applications. Additionally, providing extensive guidance on developing efficient, clinician-driven machine learning models can contribute to the creation of more dependable and trustworthy models for medical applications.

## Data availibility
The COVID-Net Biochem works and associated scripts are available in an open source manner at here, http://bit.ly/covid-net, referred to as COVID-Net Biochem. Further inquires can be directed to the corresponding author/s.

## Author contributions
H.A., M.P. and A.W. conceived the experiments, H.A., A.H., M.P. and A.W. conducted the experiments, H.A., M.P., M.J.S., A.F., A.H. and A.W. analysed the results. All authors reviewed the manuscript.

## Competing interests
The authors declare no competing interests.

## Additional information
**Correspondence** and requests for materials should be addressed to H.A.

**Reprints and permissions information** is available at www.nature.com/reprints.

**Publisher's note** Springer Nature remains neutral with regard to jurisdictional claims in published maps and institutional affiliations.